\documentclass[]{ecai} 

\usepackage{amsmath}
\usepackage{color}
\usepackage{url}
\usepackage{hyperref}
\usepackage{cleveref}
\usepackage{xurl}
\usepackage{paralist}
\usepackage{quoting}
\usepackage[utf8]{inputenc}
\usepackage{multirow}
\usepackage{wrapfig}
\usepackage{graphicx} 
\usepackage[nolist]{acronym}
\usepackage{multirow,colortbl,hhline,tabularx}
\usepackage[normalem]{ulem}
\usepackage{caption}

\newcommand{\BibTeX}{B\kern-.05em{\sc i\kern-.025em b}\kern-.08em\TeX}

\newcommand{\quotes}[1]{``#1''}

\definecolor{darkgreen}{RGB}{0,100,0}

\begin{document}


\begin{frontmatter}


\paperid{361} 


\title{Aligning XAI with EU Regulations for Smart Biomedical Devices: A Methodology for Compliance Analysis}


\author[A]{\fnms{Francesco}~\snm{Sovrano}\orcid{0000-0002-6285-1041}\thanks{Corresponding Author. Email: francesco.sovrano@uzh.ch.}}
\author[B]{\fnms{Michael}~\snm{Lognoul}\orcid{0009-0005-5137-8278}}
\author{\fnms{Giulia}~\snm{Vilone}\orcid{0000-0002-4401-5664}}

\address[A]{University of Zurich}
\address[B]{University of Namur (CRIDS, NADI)}

\begin{abstract}
Significant investment and development have gone into integrating Artificial Intelligence (AI) in medical and healthcare applications, leading to advanced control systems in medical technology. However, the opacity of AI systems raises concerns about essential characteristics needed in such sensitive applications, like transparency and trustworthiness.
Our study addresses these concerns by investigating a process for selecting the most adequate Explainable AI (XAI) methods to comply with the explanation requirements of key EU regulations in the context of smart bioelectronics for medical devices. 
The adopted methodology starts with categorising smart devices by their control mechanisms (open-loop, closed-loop, and semi-closed-loop systems) and delving into their technology. Then, we analyse these regulations to define their explainability requirements for the various devices and related goals. Simultaneously, we classify XAI methods by their explanatory objectives. 
This allows for matching legal explainability requirements with  XAI explanatory goals and determining the suitable XAI algorithms for achieving them. 
Our findings provide a nuanced understanding of which XAI algorithms align better with EU regulations for different types of medical devices. We demonstrate this through practical case studies on different neural implants, from chronic disease management to advanced prosthetics.
This study fills a crucial gap in aligning XAI applications in bioelectronics with stringent provisions of EU regulations. It provides a practical framework for developers and researchers, ensuring their AI innovations advance healthcare technology and adhere to legal and ethical standards.
\end{abstract}

\end{frontmatter}

\begin{acronym}
    \acro{AI}{Artificial Intelligence}
    \acro{XAI}{Explainable Artificial Intelligence}
    \acro{GDPR}{General Data Protection Regulation}
    \acro{LEG}{Legal Explanatory Goal}
    \acro{MDR}{Medical Devices Regulation}
    \acro{ML}{Machine Learning}
    \acro{AIA}{Artificial Intelligence Act}
    \acro{EU}{European Union}
    \acro{EDPB}{European Data Protection Board}
    \acro{DNN}{Deep Neural Network}
\end{acronym}

\section{Introduction} \label{sec:introduction}

The 2023 \ac{AI} Index Report by Stanford University reveals that medical and healthcare applications represent one of the largest investment areas in \ac{AI} (nearly 6 billion USD).
Incorporating \ac{AI} into smart bioelectronics for medical devices represents a tremendous leap in medical technology.
This integration has resulted in substantial improvements in patient care, primarily by developing advanced control systems that can adapt in real-time to patient needs, thereby greatly enhancing the effectiveness of treatments and quality of life \cite{guger2024brain}.
A key evolution in medical technology is the shift from open-loop systems, where physicians interpret data to inform decisions, to more sophisticated closed-loop and semi-closed-loop systems, where devices autonomously or semi-autonomously adjust their operations based on continuous monitoring.

\begin{figure}[!htb]
    \caption{
        Schematic overview of our research methodology for integrating legal requirements and XAI tools (cf. Section \ref{sec:methodology}).
    } \label{fig:xai_law_integration}
    \centering
    \includegraphics[width=.9\columnwidth]{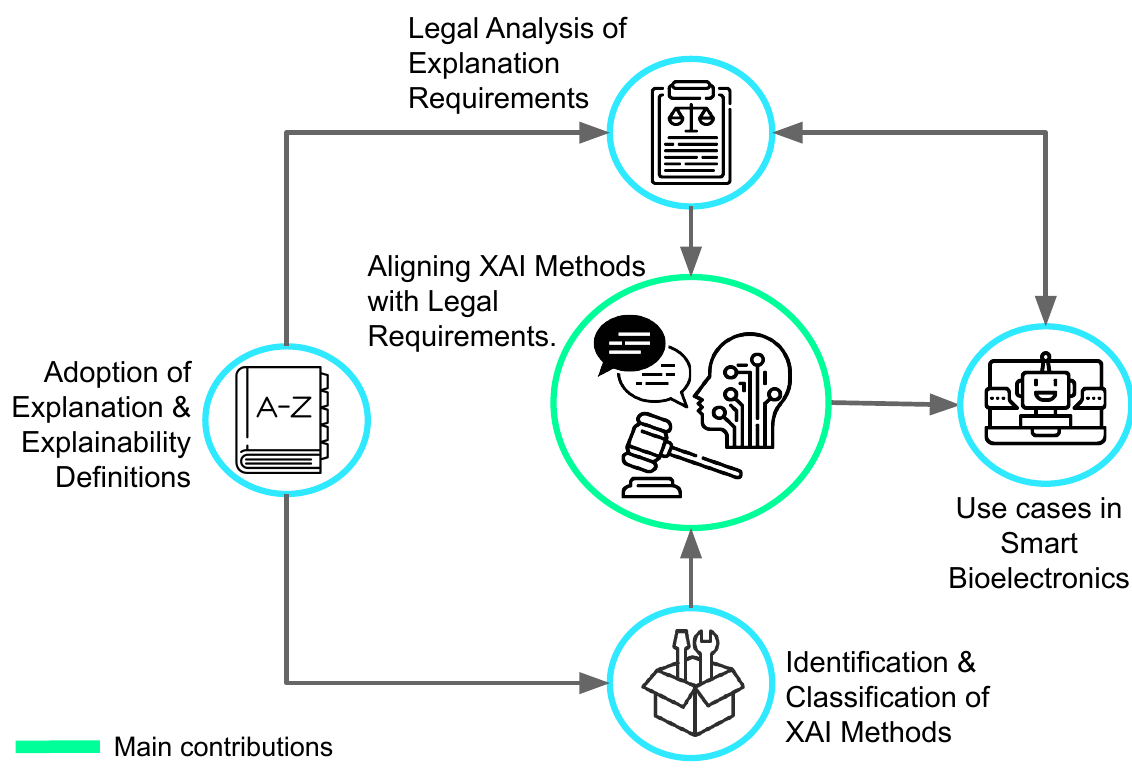}
    
\end{figure}

A significant challenge with advanced \ac{AI} systems is their `black-box' nature, which makes it hard to understand how they make decisions \cite{bernal2022transparency}. 
This challenge is critical in healthcare, where \ac{AI} systems must be accurate, transparent, and accountable to enhance trust in their users and enforce responsibility \cite{bernal2022transparency}. 
This is where \ac{XAI} plays a crucial role, offering tools to make the inner workings of these complex systems more understandable to the diverse stakeholders involved in their operation \cite{sovrano2022explanatory}. 
Regulatory frameworks, particularly in the EU, implicitly require XAI to ensure AI technologies' transparency, fairness, and accountability, as emphasised in various scholarly works \cite{sovrano2022survey}. 

EU regulations are especially stringent regarding smart bioelectronics for medical devices. These devices must comply with the \ac{GDPR}~\cite{europeancommission2016gdpr}, the \ac{AIA}~\cite{europeancommission2024aiact}, and the \ac{MDR}~\cite{europeancommission2017mdr}, each contributing unique requirements related to explainability.
However, navigating the complex regulatory landscape poses significant challenges for developers and researchers. Implementing XAI algorithms in line with EU regulations is a major hurdle, accentuating a disconnect between theory and practice in this field \cite{richmond2024explainable,kirat2022fairness}.
The motivation for our study stems from this very challenge. 
We carried out a thorough analysis of various XAI algorithms to determine if they can help satisfy explainability requirements set by the \ac{GDPR}, the \ac{AIA}, and the \ac{MDR}. 
To this end, we have developed a novel methodology, summarised by Figure \ref{fig:xai_law_integration}, to evaluate and understand the role of XAI in adhering to this legal framework. 
This study advances current understanding by categorising XAI algorithms based on their explanatory goals and matching them to the goals pursued by explainability requirements, guiding developers and researchers in selecting XAI algorithms for bioelectronic devices that better comply with EU regulations.

\section{Related Work} \label{sec:related_work} 

Integrating XAI tools into compliance processes to match explanation requirements, as contained in diverse fields of law, is still an open, multi-faceted and multi-disciplinary challenge. One of the significant stumbling blocks discussed by \citet{richmond2024explainable} is harmonising the logic followed by AI algorithms with legal reasoning and legal requirements to provide reasons or explanations. 





Previous works delved into the legal and ethical requirements for explainability in \ac{ML} \cite{bibal2021legal}, particularly in the context of the \ac{GDPR} \cite{ebers2020regulating}, the \ac{AIA} \cite{sovrano2022survey}, or other EU regulations related to sensitive fields such as finance~\cite{vilone2024explainability}. However, our study stands out in its comprehensive and practical approach, as these different studies are not oriented towards offering a methodology to find the right XAI tools.
%
\citet{bibal2021legal} investigated the increasing legal requirements for AI explainability in private and public decision-making contexts. They emphasised the implementation of these requirements in \ac{ML} models and advocated for interdisciplinary research in explainability. We went one step further by providing the kind of interdisciplinary research and methodology they suggested.

Similarly to \citet{DBLP:journals/corr/abs-2302-03180}, which proposes a strategy for choosing the proper \ac{XAI} method for specific goals, we reviewed the \ac{XAI} research to provide a synopsis of recent \ac{XAI} methods and their characteristics that make them suitable candidates for healthcare.
In contrast, our research carried forward by explicitly mapping \ac{XAI} methods to the regulatory requirements of the \ac{GDPR}, \ac{AIA}, and \ac{MDR} in the context of smart bioelectronics for medical devices. Our work delves into the obligations discussed at a general level by \citet{bibal2021legal} and translates them into operational and practical terms through focused case studies.

Although there are currently no explicit mandates for the use of \ac{XAI} systems, as noted by \citet{ebers2020regulating} in their analyses of the \ac{GDPR}, our research echoes \citet{schneeberger2020european} in emphasising the crucial role of state-of-the-art \ac{XAI} for ensuring compliance with various legal texts applicable in the medical sector, for instance, in protection of patient's sensitive data.
However, differently from \citet{schneeberger2020european}, which provides an overview of the EU's legal approach to \ac{AI} in the medical sector, our study goes beyond the general legislative landscape to perform a detailed analysis of specific \ac{XAI} methods and their potential use for compliance with these regulations, focusing on applications for smart biomedical devices. 

Additionally, our study distinguishes itself from \citet{DBLP:conf/icail/GorskiR21} by focusing on the medical field, using a broader array of \ac{XAI} algorithms and systems, and conducting an extensive qualitative analysis of legal requirements, unlike their focus on the accuracy of explainability methods like Grad-CAM, LIME, and SHAP in legal text classifications as assessed by legal professionals. 

\section{Methodology} \label{sec:methodology}

This research aims to bridge the gap between the technical capabilities of \ac{XAI} and the legal requirements set forth by key EU regulations within the domain of smart bioelectronics for medical devices: \ac{GDPR}, \ac{AI} Act, and \ac{MDR}. Our multifaceted methodology combines legal analysis with technical assessment and classification of \ac{XAI} algorithms to increase regulatory compliance. As shown in Figure \ref{fig:xai_law_integration}, our methodology comprises the following steps.

\textbf{Adoption of Explanation and Explainability Definitions.}
To map \ac{XAI} methods with legal explanatory requirements, we needed to select an appropriate definition of \textit{explanation}. We adopted the definition formalised by \citeauthor{sovrano2023objective} \cite{sovrano2023objective}, which conceptualises explanations as answers to questions that produce understanding. Among the five main definitions in contemporary philosophy, this one, rooted in Ordinary Language Philosophy, is found to align best with the legal interpretation of explanations \cite{DBLP:conf/xai/SovranoV23,sovrano2022survey}. According to this definition, an explanation provides sufficient information for an audience's understanding. This differs from other definitions that require an explanation tailored to someone's mental model or showing causal relationships. Indeed, in the legal context, explanations do not necessarily need to be fully personalised \cite{wachter2017right} and can encompass more than just causality \cite{bibal2021legal,sovrano2020modelling}.

\textbf{Legal Analysis of Explanation Requirements.}
A legal expert (the 2nd author of this paper) thoroughly analysed the \ac{GDPR}, \ac{AI} Act, and \ac{MDR} to pinpoint their explanation requirements and characteristics. Then, following an inductive coding approach \cite{fereday2006demonstrating}, the legal expert identified the high-level explanatory goals underlying these requirements, e.g., ensuring that systems' deployers understand risks related to the use of an AI system or can interpret a system's output, guaranteeing that outputs can be reviewed or contested, etc.

\textbf{Identification and Classification of \ac{XAI} Methods.}
Concurrently, two \ac{AI} experts (the 1st and 3rd author of this paper) conducted in three phases a literature review to compile a comprehensive (but not exhaustive) list of existing \ac{XAI} methods. Initially, the search query \textit{\quotes{XAI survey}} was used on Google Scholar, targeting relevant publications in top journals from 2023. 
Subsequently, the research scope was broadened to incorporate insights from the XAI survey of \citet{vilone2021classification}, to ensure a more comprehensive synopsis. 
Finally, the list of XAI algorithms was integrated with algorithms known to the experts but not mentioned in the surveyed literature.
These algorithms were categorised based on their explanation format, input format, and model-agnostic status to discern the types of explanatory questions they could address, such as \quotes{what happens if feature X is changed} or \quotes{what is the contribution of feature Y to the output}. We used a question-driven design process similar to that of \citet{liao2021question}, in which \ac{XAI} methods are mapped to explanatory questions based on their characteristics. Differently from \citet{liao2021question}, our mapping did not involve only interrogative particles (e.g., why, how), but we formulated complete questions (see Table \ref{tab:model_agnostic_xai_questions}) via an inductive coding approach \cite{fereday2006demonstrating}, allowing the questions to emerge naturally from the characteristics of the \ac{XAI} methods. 

\textbf{Aligning \ac{XAI} Methods with Legal Requirements.}
By performing a deductive thematic analysis \cite{fereday2006demonstrating}, we mapped the \ac{XAI} questions to the legal explanatory goals enshrined in the \ac{GDPR}, \ac{AIA}, and \ac{MDR}. This was possible because the adopted definition of explanation is framing explanations as answers to questions.
Eventually, we could identify congruence where \ac{XAI} capabilities can be exploited to help meet the stipulated legal explanation requirements.
This matching process ensures that the selection of \ac{XAI} methods is technologically sound and legally robust. 
To aid developers and researchers in selecting the most appropriate \ac{XAI} algorithms for different bioelectronic devices, we developed a set of instructions (see Section \ref{sec:discussion}). These instructions and the methodology provide a fundamental framework designed to be flexible and seamlessly incorporate newly emerging \ac{XAI} algorithms and evolving regulations.

\textbf{Identification of Case Studies in Smart Bioelectronics.}
We focused on two use cases in the field of smart bioelectronics determined by the type of control they employed: 1) closed-loop and 2) open- and semi-closed-loop. The distinction is significant as it influences the decision-making processes and the applicable legal frameworks. For example, the \ac{GDPR}'s right to explanation pertains to high-stakes, fully automated algorithmic decision-making in closed-loop systems.

\section{Background} \label{sec:background}

This section provides background information on AI-based biomedical technologies, EU regulations, and \ac{XAI}.


\subsection{Smart Bioelectronics and Biomedical Devices} \label{sec:background:medicine}

%
\textit{Biomedical devices} is an umbrella name that covers a wide variety of tools used to
help diagnose, prevent, and treat diseases \cite{lam2019biomedical}.
\textit{Bioelectronics} refers to a subset of specialised biomedical devices that combine electronic technology, like sensors, with biology and medicine. These devices can interact with biological systems, from whole organs to tiny cellular components, in various ways, such as using light, magnetic, or chemical methods \cite{jafari2023merging}. 
Based on their decision-making mechanisms, bioelectronics and biomedical devices can be categorised into open-loop, closed-loop, and semi-closed-loop control systems. 
\textit{Open-loop control systems}, such as Electrocardiograms (ECG), provide only outputs instrumental in the decision-making of healthcare professionals. 
%
In contrast, \textit{closed-loop systems} autonomously adjust their operations based on continuous monitoring. For example, artificial pancreas systems for diabetes management autonomously monitor glucose levels and administer insulin \cite{lin2019strategies}.
\textit{Semi-closed-loop systems} represent an intermediate approach where the machine instructs a patient to manually perform life-saving actions based on data (e.g., manual insulin injection adjustments based on a Continuous Glucose Monitoring System \cite{lin2019strategies}). 

In addition to loop-based categorisation, biomedical devices can be classified based on their potential risks to human health. This classification is influenced by the device's operating mode and characteristics, such as whether it is invasive or non-invasive. The classification is determined with relevant legislation (i.e., \ac{MDR})~\cite{almpani2022computational}.

Neural implants represent a fascinating intersection of \ac{AI} and neurotechnology and can be split between Brain-Computer Interfaces (BCIs) and Computer-Brain Interfaces (CBIs). Both prosthetic devices establish direct communication between the human brain and external hardware or software. BCIs use decoding algorithms to restore lost functions, while CBIs exploit encoding algorithms to convert external sensory signals to neural stimulation patterns~\cite{rao2023brain}.
Depending on their level of autonomy and decision-making mechanisms, neural implants are subject to different explanation requirements (see Table \ref{tab:legal_requirements}).
One example of a closed-loop neural implant is the Responsive Neuro Stimulation (RNS) system, which is designed for individuals with epilepsy who do not respond well to medications and are not candidates for epilepsy surgery. Epileptic seizures are caused by abnormal electrical activity in the brain. RNS system records intracranial EEG patterns to timely activate a stimulation designed to mitigate such activity~\cite{gouveia2024neurostimulation}.
A Spinal Cord Stimulator (SCS) is instead an example of semi-closed loop neural implant used to alleviate chronic pain. It consists of an implanted device that delivers electrical pulses to the spinal cord to disrupt pain signals before they reach the brain. Unlike closed-loop systems, where all adjustments are fully automated, patients and healthcare professionals often have important control over these devices and their stimulation decisions. They can adjust the stimulation settings within certain limits, such as changing the intensity, frequency, or coverage of the pulses~\cite{garcia2020spinal}.

\subsection{EU Regulations Relevant to Smart Biomedical Devices: Scopes and Notions} \label{sec:background:law}


The \textit{\acl{MDR}}~\cite{europeancommission2017mdr} governs the placing on the market and use of medical devices in the EU (Art. 1.1) 
for the diagnosis, prevention, prediction, monitoring, treatment, 
of diseases, injuries, or disabilities. These devices must primarily operate not through pharmacological, immunological, or metabolic means but can be supported by them (Art. 2.1).

The \textit{\acl{GDPR}}~\cite{europeancommission2016gdpr} instead applies to personal data processing 
\begin{inparaenum}[\itshape i)\upshape]
\item by EU-based data controllers or processors, or 
\item involving EU residents' data processed by a controller located outside the EU, 
(Art. 2 and 3).
\end{inparaenum}
Personal data is information about an identifiable person: the `data subject' (Art. 4.1). `Processing' encompasses: collection, organisation, storage, consultation, use, disclosure and erasure (Art. 4.2). A `data controller' sets personal data processing purposes and means (Art. 4.7), while a `processor' handles data on behalf of a controller (Art. 4.8).

The \textit{\acl{AIA}}~\cite{europeancommission2024aiact}, adopted in June 2024, applies to AI systems marketed or used in the EU or whose outputs are employed in the EU, regardless of the provider's or deployer's location (Art. 2). The AI systems covered are software able to infer, from their inputs, how to generate outputs (e.g., predictions, content, recommendations, or decisions, Art. 3.1). 
A `provider' develops or commissions AI systems for market placement or service (Art. 3.2), and a `deployer' employs an AI system, excluding for personal, non-professional use (Art. 3.4). 
High-risk AI systems include those covered by EU legislation listed in Annex I, like \ac{MDR}, when requiring third-party conformity assessments and those listed in Annex III, e.g., for remote biometric identification (Art. 6).

\subsection{XAI Algorithms} \label{sec:background:xai}

\ac{XAI} literature features a variety of domain-dependent and context-specific methods that differ in their explanation generation strategies, formats, and applicability to disparate data and learning algorithms \cite{islam2022systematic}. 
Researchers have developed taxonomies to aid in selecting suitable \ac{XAI} methods for specific problems.
A key contribution to this paper comes from the work of \citet{liao2020questioning}, who categorise \ac{XAI} methods based on the questions they address, and ~\cite{vilone2021classification}, who organised \ac{XAI} methods by stage, scope, and format.
Firstly, the stage category splits explanations between ante-hoc and post-hoc. Ante-hoc methods aim to build inherently explainable models, while post-hoc methods seek to clarify the logic of an already trained model using an external explainer. Secondly, explanations are divided between having a global (explaining the entire model's process) and a local (explaining individual inferences) scope \cite{vilone2021classification}.
Thirdly, explanations differ in their format. Some consist of vectors, tensors or matrices of numbers pointing out the most relevant input features. Other explanatory formats are texts, charts and diagrams, rules, or a combination of these formats.
%
As discussed in Section \ref{sec:expl_req_and_LEG} and shown in Table \ref{tab:legal_goals}, some regulations favour ex-ante explainability \cite{rudin2019stop}. However, much of the research in XAI focuses on post-hoc solutions \cite{vilone2021classification}. 


\section{Explanation Requirements and Legal Explanatory Goals} \label{sec:expl_req_and_LEG}

This section examines the EU regulations outlined in Section \ref{sec:background:law}, focusing on their mandates for explanations. Our analysis encompasses the contents and formats of the explanations required.
We also note the interactions and overlaps among rules imposed on devices with varying autonomy. Next, we delineate and classify the legal objectives derived from these requirements. This offers a holistic view of the goals behind the explanation requirements in the EU's regulatory framework for AI and digital health technologies.


\textbf{\acl{MDR}.} The \ac{MDR} (Art. 10.11 and Annex I) mandates that medical devices include user instruction that, aimed at users or patients, must cover:
\begin{inparaenum}[\itshape i)\upshape]
	\item the device's intended purpose;
	\item indications, contra-indications, residual risks, and side effects;
	\item target patient groups; 
    \item performance characteristics;
	\item suitability information for healthcare professionals;
	\item user requirements for proper device usage;
	\item any preparatory treatment needs, like calibration; and
	\item guidance on verifying device installation and operational readiness (Annex I.23.4).
\end{inparaenum}

Art. 2.37 of \ac{MDR} distinguishes between `patient' and `user' (healthcare professional or layperson), affecting the complexity of explanations based on the intended audience. For healthcare professionals, detailed instructions are suitable, while simpler information and language are necessary for laypersons or patients. Regardless of the audience, the \ac{MDR} mandates \quotes{easily legible and comprehensible} instructions. The \ac{MDR} does not specify a format but suggests written explanations, allowing for graphical and numerical forms.
The \ac{MDR}'s primary goals are to enhance transparency and safety around medical devices for public health and to empower users and patients to make informed decisions (recital 43).

\textbf{\acl{GDPR}.} The \ac{GDPR} regulates decisions made solely through automated means and involving the processing of personal data, with legal or significant effects on individuals (Art. 22.1). It imposes safeguards, including providing data subjects with \quotes{meaningful information about the logic involved} and the significance and consequences of processing (Art. 13.2(f), 14.2(g), 15.1(f)). Additionally, Recital 71 mentions the obtaining of \quotes{an explanation of the decision reached} to enable individuals to understand and contest decisions (Art. 22, Recital 71) and potentially influence future behaviour to obtain a desired outcome, though this aspect is less emphasised (e.g., see p. 26 \cite{EDPBGuidelines}). The \ac{EDPB} advises data controllers to explain \quotes{the rationale behind, or the criteria relied on} for these decisions (p. 25 \cite{EDPBGuidelines}).
%
The \ac{GDPR}'s key principles include transparency in data processing and empowering data subjects (Art. 5, 12-22; Recitals 29, 58, 60). Therefore, it requires explanations to be adapted to the recipients' background knowledge \cite{Meaningfulinformation} and to be \quotes{concise, transparent, intelligible, and easily accessible, using clear and plain language} (Art. 12.1). The \ac{EDPB} further clarifies that explanations should enable understanding of the reasons behind decisions without disclosing complex algorithmic details (p. 25) \cite{EDPBGuidelines}. Both the \ac{GDPR} and \ac{EDPB} guidelines do not prescribe a specific format for explanations, allowing for flexibility. 

\textbf{\acl{AIA}.} The \ac{AIA} sets strict transparency standards for high-risk AI systems, demanding them to be \quotes{sufficiently transparent to enable deployers to interpret and use them appropriately} (Art. 13.1) and to include detailed instructions for use (Art. 13.2), covering:
\begin{inparaenum}[\itshape i)\upshape]
    \item system characteristics, capabilities, and limitations of performance, including: 
    \begin{inparadesc}
        \item its intended purpose,
        \item accuracy (with metrics), robustness, and cybersecurity,
        \item potential health, safety, or fundamental rights concerns,
        \item when available, ability to provide information explaining its output,
         \item when appropriate, performance characteristics for target groups,
        \item when appropriate, specifications about input data and information on training, validation, or testing datasets, 
         \item when available, information enabling deployers to interpret the system's output and use it appropriately; 
    \end{inparadesc}
    \item any planned changes to the system or its performance;
    \item human oversight measures, including the measures to facilitate system outputs interpretation; 
    \item the system's expected lifetime and maintenance requirements, as well as the resources needed to run it; and
    \item a description of the mechanisms to collect, store and interpret the system's logs.
\end{inparaenum}

The \ac{AIA} also mandates human oversight measures (Art. 14), requiring instructions for deployers and human oversight personnel (Art. 13, 14) with sufficient information for understanding system capacities and limitations, interpreting outputs, making usage decisions, and intervening if necessary.
These instructions should be \quotes{concise, complete, correct, clear, relevant, accessible, and comprehensible} (Art. 13.2). The \ac{AIA} also emphasises the necessary competence, training, and authority needed for oversight personnel (Art. 26), suggesting detailed explanations are essential.
Explanations may be in various formats, including digital (Art. 13), potentially using text, visuals, or interactive tools (Art. 14). The \ac{AIA} aims to ensure that deployers can understand and properly use high-risk systems and maintain operator control (Recitals 72, 73), supporting a \quotes{high level of protection of health, safety, and fundamental rights} (Recital 1).

\textbf{Legal Explanatory Goals.} The explanation requirements detailed so far apply to any bioelectronic component and biomedical device that enters the scope of \ac{GDPR}, \ac{AIA}, and \ac{MDR} and meets the conditions which trigger their explanation requirements (e.g., fully automated high-stakes decisions based on personal data for the \ac{GDPR}). These requirements are not mutually exclusive, and a cumulative application of two (or more) requirements may be needed. In particular, the device's autonomy level will influence the number of requirements to be complied with, as illustrated in Table \ref{tab:legal_requirements}.  

\begin{table*}
    \caption{Applicability of Legal Requirements to Different Device Types.}
    \label{tab:legal_requirements}
    \centering
    \resizebox{\linewidth}{!}{
        \begin{tabular}{|l|l|l|l|}
            \hline
            \textbf{Device Type} & \textbf{MDR (Art. 10.11 and Annex I.23.4)} & \textbf{AIA (Art. 13-14)} & \textbf{GDPR (Art. 13-15 and 22)} \\ \hline
            Open-loop & Applicable (if medical device) & Applicable (if high-risk AI system) & Not applicable (no fully-automated decision) \\ \hline
            Semi-closed-loop & Applicable (if medical device) & Applicable (if high-risk AI system) & Not applicable (no fully-automated decision) \\ \hline
            Closed-loop & Applicable (if medical device) & Applicable (if high-risk AI system) & Applicable (if high-stakes decision) \\ \hline
        \end{tabular}
    }
\end{table*}

Based on the analysis of the legal explanatory requirements led in this section and following the methodology described in Section \ref{sec:methodology}, we identified 11 high-level legal explanatory goals to which these requirements pertain. 
Table ~\ref{tab:legal_goals} presents the identified goals, specifying the relationship with the EU regulations. Importantly, we noted that goals and regulations have a many-to-many relationship, as each goal may be related to one or more regulations. On top of that, building on the notions described in \ref{sec:background:xai}, we clarify in Table ~\ref{tab:legal_goals} whether each goal requires global or local explanations to be achieved, as well as the stage at which they should be provided: ex-ante, or ex-post.


\begin{table*}
    \caption{Legal Explanatory Goals, Related Regulations, and their Scope (Global/Local) and Stage (Ex-Post/Ex-Ante).}
    \label{tab:legal_goals}
    \centering
        \begin{tabular}{|c|l|l|l|l|}
        \hline
        \textbf{ID} & \textbf{Legal Explanatory Goal} & \textbf{Related Regulation(s)} & \textbf{Scope} & \textbf{Stage} \\ \hline
        A & Understand the risks related to the use of the system & MDR, AIA & Global & Ex-ante \\ \hline
        B & Understand the conditions under which the intended users should use the system or opt-out & MDR, AIA & Global & Ex-ante \\ \hline
        C & Understand the consequences of decisions taken by the system & GDPR & Any & Any \\ \hline
        D & Ensure that decisions taken by the system can be reviewed or contested & GDPR, AIA & Local & Ex-post \\ \hline
        E & Understand what to do to change a future decision of the system & GDPR & Any & Any \\ \hline
        F & Detect and address anomalies, dysfunctions, or unexpected performance & AIA, MDR & Any & Any \\ \hline
        G & Understand why a specific decision has been taken & GDPR, AIA & Local & Ex-post \\ \hline
        H & Understand how to use the system & MDR, AIA & Global & Ex-ante \\ \hline
        I & Understand the general logic of the system & AIA & Global & Ex-ante \\ \hline
        J & Understand the accuracy scores and the performance of the system’s outputs & AIA, MDR & Global & Ex-ante \\ \hline
        K & Interpret the system’s output & AIA & Local & Ex-post \\ \hline
        \end{tabular}
\end{table*}

\section{A Categorisation of XAI in Terms of Explanatory Goals}
\label{sec:xai_explanatory_goals}

This section presents a categorisation of \ac{XAI} methods identified in the XAI literature and elaborates on their roles in fulfilling the legal explanatory goals of Section \ref{sec:expl_req_and_LEG}. This classification stems from the methodology outlined in Section \ref{sec:methodology}, considering that XAI explanations answer specific questions about AI models and their outputs.

%
We organised the XAI methods based on explanation format, scope, input type, stage of application, and model specificity. This classification, grounded on established taxonomies (cf. Section \ref{sec:background:xai}) and \citet{liao2020questioning}'s methodology, 
aids in pinpointing which explanatory question can be answered by each XAI method and, subsequently, which explanatory goals it addresses. 
\citet{liao2020questioning} exploited the explanation format to identify which questions can be answered by the XAI methods. For example, counterfactual methods \cite{he2023counterfactual} inspect how the output changes when the input instance is modified, generating a `what-if' scenario that manifests what leads to a desired outcome. 
Thus, these explanations can answer the question \quotes{What minimal changes would need to be made to input to change its prediction?}.
Instead, similarity-based XAI methods \cite{poche2023natural} show how the model behaved with similar inputs, addressing questions like \quotes{Which past instances yielded similar predictions to this input?}.

Our methodology for aligning XAI questions with legal explanatory goals (Tables \ref{tab:model_specific_xai_questions}, \ref{tab:time_series_xai_questions}, and \ref{tab:model_agnostic_xai_questions}), as detailed in Section \ref{sec:methodology}, involves evaluating each XAI question against regulatory requirements.
Specifically, we matched legal explanatory goals requiring global explanations to global XAI methods only, while legal goals with a local perspective were linked to local or global XAI methods, as global explanations can sometimes provide local insights.
Global feature attribution XAI methods like TreeSHAP \cite{lundberg2020local} are suitable to respond to the question \quotes{What are the most important features influencing all the model’s predictions?}, but not \quotes{How does a specific feature influence the prediction for an individual instance?}, which has a more local scope and can be addressed with XAI methods such as LIME \cite{ribeiro2016model}.  
Indeed, TreeSHAP's focus is on understanding the slightest changes in the input features that would lead to a different prediction and showing how alternative outcomes could be achieved.
Therefore, TreeSHAP best aligns with goals A, B, F, H, and I. In general, global feature attribution XAI methods aid in achieving goals B and F as they allow intended users to make informed decisions, such as detecting anomalies and opting out of using the system if they cannot provide these features.
Goal H emphasises understanding system usage beyond following user manuals. It can involve experimenting with the system to determine the correct inputs for the desired outcome, focusing on clarity about the inputs to use. Knowing which features have more impact on the output can speed up the process of finding this sweet spot. 
Thus, it was associated with global feature attribution and rule-based methods, like Shapley Flow~\cite{wang2021shapley}.
%

Instead, local model-agnostic XAI methods, like the counterfactual and contrastive explainers, can identify minimal input changes for different outcomes, aiding in goals D, E, F, G, and K. Yet, it cannot clarify when/how to use the system or the logic underlying its predictions.
%
%
%
Not all local XAI methods can answer those five goals. For instance, given that we interpreted goal E as needing specific instructions to alter the system's output, local feature attribution and saliency maps do not contribute to reaching that goal. They simply highlight important features without suggesting how to modify them. Often, these features cannot be eliminated merely by zeroing them out, so goal E was not linked to either approach. These two local XAI methods usually address only goals D, F, G, and K.
Also, activation maximisation and layer-wise relevance propagation provide similar explanations to feature attribution methods. The main difference is that they do not provide information to review or contest a decision (D) since they do not explain the contribution of the input's features to the output.

We also found that no XAI tool can explain the consequences of a decision; it only explains the process and reasons behind it. Thus, explaining consequences should be done manually to achieve goal C.
Similarly, we found that goal J does not normally pertain to XAI, as it is more about the usual testing and validation steps performed when building an AI system. 

For goal A, relevant risks under the considered laws are harms caused when the system works as intended (e.g., discriminatory yet accurate predictions stemming from biased data) or malfunctions. For this reason, we could only associate global feature attribution and rule-extraction XAI algorithms with A.


The stage of an XAI method (ex-ante or ex-post) also influences the question framing. 
Ex-ante methods help understand the data and its features before a prediction is made or finalised, leading to questions like \quotes{What set of rules does the model follow to make all predictions?}. Such questions are associated with legal goals A, B, H-J.
Instead, ex-post methods drive questions about interpreting these predictions, such as \quotes{Why did the model make this specific prediction for this instance?}, which are linked to goals D, G and K.

The XAI methods can be segmented into model-specific and model-agnostic, as systematically presented in Tables~\ref{tab:model_specific_xai_questions}-\ref{tab:model_agnostic_xai_questions}. This arrangement facilitates the identification of appropriate XAI methods for case studies in smart bioelectronics for medical devices (see Section \ref{sec:case_studies}).
%
Model-specific XAI methods, listed in  Table \ref{tab:model_specific_xai_questions} alongside the questions and legal goals we mapped them with, are tailored to specific model types, such as tree-based models or \acp{DNN}.  
Within this category, we decided to show separately, in Table \ref{tab:time_series_xai_questions}, those methods designed for time series AI models.
They are particularly relevant to biomedical devices because they must work with data sequences that vary over time, such as the heart rate in the case of ECGs.
Temporal integrated gradients \cite{duell2023formal}, for instance, help identify patterns and segments in time series that are deemed critical by a model to interpret outputs from devices like pacemakers or ECGs. 
Conversely, model-agnostic XAI algorithms (Table \ref{tab:model_agnostic_xai_questions})
apply to any model type, providing flexibility in their application. For instance, LIME \cite{ribeiro2016model}, SHAP and KernelSHAP \cite{lundberg2017unified} identify the most important features influencing predictions in any AI model. 

\begin{table}[t]
    \scriptsize
    \caption{Model-Specific XAI Methods with Explanatory Goals.} \label{tab:model_specific_xai_questions}
    \centering
    \resizebox{\linewidth}{!}{
        \begin{tabular}{|p{2cm}|p{1.2cm}|p{1cm}|p{0.5cm}|p{2.3cm}|}
            \hline
            \textbf{Question} & \textbf{XAI \newline Algorithms} & \textbf{Applicable Model Types} & \textbf{Expl. Goal ID(s)} & \textbf{Explanation} \\ \hline
            What are the most important features influencing \underline{all} \underline{predictions} of the model? & Global feature attribution methods: TreeExplainer~\cite{lundberg2020local}, CAVs \cite{kim2018interpretability} & Tree-based models, \acp{DNN} & A, B, F, H, I & Identify key features for predictions to understand system's conditions of use, risks, general logic, and detect anomalies.  \\ \hline
            How do interactions between features affect \underline{all} \underline{predictions} of the model? & Global feature attribution methods \cite{lundberg2020local} & Tree-based models, Bayesian networks & A, B, F, H, I & Identify key features for predictions to understand system's conditions of use, risks and general logic and detect anomalies. \\ \hline
            What is the \underline{model's} \underline{inner} \underline{logic}? &  NNKX \cite{bondarenko2017classification}, ExtractRule \cite{fung2005rule} & \acp{DNN}, SVM & A, B, D-I, K & Transparent models mimicking the behaviour of a black-box. \\ \hline
            What set of rules does the model usually follow to make \underline{all} \underline{predictions}? & Rule-based algorithms \cite{dash2018boolean,wang2021shapley}& Decision trees, random forests, linear models, \acp{DNN} & A, B, D-I, K & Clarify the rules to understand system usage, risks, logic, decisions, outputs, how to contest or change them, and detect anomalies. \\ \hline
            What parts of an input (e.g., an image) influence the model's decision? & Saliency maps \cite{feldhus2022constructing}, LRP \cite{bach2015pixel} & \acp{DNN} & D, F, G, K & Identify influential input areas to review or contest decisions, detect anomalies, interpret specific decisions and outputs. \\ \hline
            What contributions do individual neurons in a DNN make to the final prediction? & Activation maximisation \cite{li2021deep} & \acp{DNN} & F, G, K & Analyse neuron contributions for anomaly detection, decision and output interpretation. \\ \hline
            How do different layers in a \ac{DNN} contribute to a prediction? & Layer-wise relevance propagation \cite{li2021deep, kim2018interpretability} & \acp{DNN} & F, G, K & Examine layer contributions for anomaly detection, decision and output interpretation. \\ \hline
            How to interpret a neural net’s internal state in terms of human-friendly concepts? & Concept-based methods \cite{kim2018interpretability} & \acp{DNN} & D, F, G, K & Evaluate key contributions to decisions and outputs, review or challenge them, and identify anomalies. \\ \hline
            What are the most similar instances to a given input with respect to a model's prediction? & Self-Organising Maps (SOM)~\cite{hamel2006visualization} & SVM & D, F, G, K & Compare similar instances to review/contest/interpret decisions/specific outputs and detect anomalies. \\ \hline
        \end{tabular}
    }
\end{table}

\begin{table}[tb]
    \scriptsize
    \caption{XAI Methods for Time Series Models (neural networks) with Explanatory Goals.} \label{tab:time_series_xai_questions}
    \centering
        \begin{tabular}{|p{2.3cm}|p{1.4cm}|p{0.5cm}|p{2.8cm}|}
        \hline
        \textbf{Question} & \textbf{XAI \newline Algorithms} & \textbf{Expl. Goal ID(s)} & \textbf{Explanation} \\ 
        \hline
        What points in the time series are most important for the model's decision? & Feature \newline attribution \newline methods \cite{duell2023formal} & D, F, G, K & Identify key points to review/contest/interpret decisions/specific outputs, and detect anomalies.  \\ \hline
        What are the key segments in a time series that influence the model's output? & Saliency maps \cite{aydemir2023tempsal} & D, F, G, K & Clarify relevant segments to review/contest/interpret decisions/specific outputs and detect anomalies.  \\ \hline
        What minimal changes in a time series would alter its predicted outcome? & Counterfactual explanations \cite{he2023counterfactual} & D, E, F, G, K & Identify minimal changes leading to review/contest/interpret decisions/specific outputs, detect anomalies, and make changes to future decisions.  \\ \hline
        What parts of an input (e.g., an image) influence the model's decision? & Visual \newline attribution methods \cite{parvatharaju2021learning} & D, F, G, K & Determine influential input parts to review/contest/interpret decisions/specific outputs and detect anomalies.  \\ \hline
        How does a specific feature influence the prediction for an individual instance? & Feature importance analysis \cite{meng2023initial}& D, F, G, K & Assess feature influence on individual predictions to review/contest/interpret decisions/specific outputs and detect anomalies.  \\ \hline
        \end{tabular}
\end{table}
\begin{table}[tb]
    \scriptsize
    \caption{Model-Agnostic XAI Methods with Explanatory Goals.} \label{tab:model_agnostic_xai_questions}
    \centering
        \begin{tabular}{|p{2.3cm}|p{1.4cm}|p{0.5cm}|p{2.8cm}|}
            \hline
            \textbf{Question} & \textbf{XAI \newline Algorithms} & \textbf{Expl. Goal ID(s)} & \textbf{Explanation} \\ \hline
            What is the \underline{inner} \underline{logic} of the model? & Surrogate models~\cite{bastani2017interpretability} & A, B, D-I, K & Transparent models mimicking the behaviour of a black-box. \\ \hline
            How does a specific feature influence the prediction for an individual instance? & LIME~\cite{ribeiro2016model}, SHAP~\cite{lundberg2017unified} & D, F, G, K & Assess feature impact on individual outputs to review/contest/interpret decisions/specific output and detect anomalies. \\ \hline
            What are the most important features influencing \underline{all} \underline{predictions} of the model? & Global feature attribution methods like SHAP~\cite{lundberg2017unified} & A, B, F, H, I & Identify key features for predictions to understand the system's conditions of use, risks and general logic and detect anomalies. \\ \hline
            What are the most similar instances to a given input with respect to a model's prediction? & Similarity-based methods \cite{poche2023natural} & D, F, G, K & Compare similar instances to review/contest/interpret decisions/specific outputs and detect anomalies. \\ \hline
            What minimal changes would need to be made to an input to change its prediction? & Counterfactual explanations \cite{guidotti2022counterfactual} & D, E, F, G, K & Identify changes for different outcomes pertinent to review/contest/interpret decisions/specific outputs, detect anomalies, and make changes to future decisions. \\ \hline
            Why this output instead of another? & Contrastive explanations \cite{dhurandhar2018explanations} & D, E, F, G, K & Clarify reasoning behind outputs to review/contest/interpret decisions/specific outputs, detect anomalies, and make changes to future decisions. \\ \hline
            What are the conditions or features of the input that, when held fixed, are most responsible for a particular model's prediction or classification? & Anchors \cite{ribeiro2018anchors} & D, F, G, K & Highlights key features to review/contest/interpret decisions/specific outputs and detect anomalies. \\ \hline
            How would changing multiple features simultaneously affect the model's prediction for a specific instance? & Counterfactual and interaction detection methods \cite{guidotti2022counterfactual} & D, E, F, G, K & Examines combined feature effects relevant to review/contest/interpret decisions/specific outputs, detect anomalies, and make changes to future decisions. \\ \hline
            How can we understand the model's decision for a specific instance in the context of its training data? & Contextual analysis methods \cite{poche2023natural} & D, F, G, K & Provide decision context to review/contest/interpret decisions/specific outputs and detect anomalies. \\ \hline
        \end{tabular}
\end{table}

\section{Case Studies: Closed-Loop and Semi-Closed-Loop Control} \label{sec:case_studies}

AI-enhanced neural implants can detect early signs of stroke, improve memory, and help control paralysed limbs to perform fine motor tasks, e.g., holding a glass (cf. Section \ref{sec:background:medicine}).

Responsive Neuro Stimulations (RNS) are closed-loop systems. They are, in principle, subject to \ac{GDPR} explanation requirements as they process (sensitive) personal data to make high-stakes, fully automated decisions. 
Indeed, a stimulation performed at the wrong time, on the wrong area of the brain or with the wrong electrical pulses might have side effects with varying severity \cite{gouveia2024neurostimulation}, including pain, discomfort, sensory disturbances, etc.
RNS systems are also subject to the explanation requirements of the \ac{MDR} and the \ac{AIA}. According to the MDR rules, they must undergo a third-party conformity assessment, making them high-risk systems as well.
Since intracranial EEG patterns are extrapolated from time series, we look at Table \ref{tab:time_series_xai_questions} for suitable XAI methods.
According to our methodology, surrogate models (as also suggested by \citet{rudin2019stop}) or a mix of counterfactual, rule-based, and global feature attribution XAI methods have high chances of meeting the \ac{GDPR} (goals D, E, G), the \ac{AIA} (goals A, B, D, F-I, K) and \ac{MDR} (goals A, B, F, H) explanation requirements. However, as explained in Section \ref{sec:discussion}, these combinations may not necessarily meet all the legal explanatory goals identified, requiring the integration of other XAI tools or human intervention.


On the other hand, Spinal Cord Stimulator (SCS) systems, which are semi-closed loop devices, are in principle not subject to \ac{GDPR}'s explanation requirements. However, as RNS, SCS systems are medical devices that can lead to complications like paresthesia, infections, epidural hematoma, nerve injury, paralysis, and even death \cite{garcia2020spinal}. Therefore, they will need to comply with the explanation requirements contained in the \ac{MDR} and the \ac{AIA}: they are high-risk AI systems, as they have to undergo a third-party conformity assessment under the \ac{MDR}.
Hence, according to our methodology, surrogate models or a mix of counterfactual, rule-based, and global feature attribution XAI methods still have a high chance of meeting the \ac{AIA} and \ac{MDR} legal explanatory goals. 

\section{Instructions for Use \& Discussion of Findings} \label{sec:discussion}

This study introduces a multi-faceted, multi-step, multi-domain methodology for aligning XAI tools with EU regulations, addressing a major gap in the field \cite{richmond2024explainable}. Matching the legal and the XAI fields presents a few challenges, such as integrating their respective technical languages and reconciling the differing explanatory goals of AI systems and EU regulations \cite{richmond2024explainable}.

Nevertheless, this methodology provides a practical approach for selecting appropriate XAI tools for new AI use cases in healthcare. It involves a multi-phase process (see Figure \ref{fig:xai_law_integration}) tailored to specific applications, ensuring the chosen XAI tools are appropriate to help meet legal explanatory requirements. Readers interested in applying this work should start from Section~\ref{sec:expl_req_and_LEG}, where requirements' main features are discussed, and the legal explanatory goals are identified, and then move to Section~\ref{sec:xai_explanatory_goals}, where the XAI algorithms are concretely mapped to the objectives pursued by explanation requirements. 

Our methodology is adaptable and can accommodate future developments in AI applications, XAI algorithms, and evolving legal requirements. For instance, interested parties should follow these steps when applying our methodology to new or different explanation requirements. First, the identification of: 
\begin{inparaenum}[\itshape i)\upshape]
\item the recipients of the explanations and their background knowledge; 
\item the level of detail required from the explanation; 
\item the format imposed on the explanation, if any;
\item the explanandum (i.e., the pieces of information) required;
\item the specific objectives pursued by the law; 
\item the types of questions that the explanation should answer; 
\item the moment at which they should be provided; 
\item the scope of the required explanation.    
\end{inparaenum}
Second, based on these major features, taking an inductive approach, interested parties could either relate the requirements studied to one or more of the high-level legal explanatory goals identified in our work or determine other explanatory goals. Finally, by contrasting the identified goals with the question(s) that specific XAI methods or algorithms are meant to answer, interested parties should find matches between both and be able to select appropriate tools to help answer explanation requirements.

Conversely, readers can start by determining their chosen explanation and input format, the model's applicability (model-specific or model-agnostic) and then integrate novel XAI algorithms into our methodology. With this foundation, they can inductively formulate explanatory questions that these algorithms are meant to address, generating complete questions. Subsequently, they can follow a similar process of matching these questions with the relevant legal explanatory goals, ensuring alignment between legal requirements and XAI capabilities.

Finally, each regulation we have examined in our research serves multiple orthogonal objectives (e.g., \ac{MDR} aligns with the legal explanatory goals A, B, F, H, J). Consequently, even if an XAI tool is designed to address one specific goal (or more), it may not fully encompass all the legal explanatory goals intended by a given regulation. Additionally, individual XAI tools frequently only partially fulfil the objectives associated with a particular goal.
Therefore, practitioners and researchers need a case-by-case assessment when applying our work in real-world scenarios. This assessment should determine the extent to which the selected XAI methods are required to implement a sufficiently diverse set of tools to ensure compliance with legal requirements while avoiding unnecessary redundancies.
%

\section{Threats to Validity} \label{sec:limitations}

\textbf{Extrinsic Threats.} Extrinsic threats include the potential for new interpretations of the Regulations discussed (e.g., through EU case-law), which may alter the applicability of our findings. Additionally, while our prescribed compliance methods assist in obtaining the necessary information, the effectiveness of conveying this information to individuals with different expertise and background knowledge is yet to be determined. Furthermore, while our study concentrates on biomedical devices and their significant requirements, it is crucial to acknowledge that other EU or national laws might impose additional explanation requirements in specific contexts. As a result, some devices may encounter extra constraints, potentially necessitating a broader range of XAI tools than those discussed in this paper.

Another extrinsic limitation arises from the inherent complexities in explainability. Most existing XAI methods, such as surrogate models, SHAP, and LIME, often rely on imperfect heuristics and usually operate effectively under specific conditions, lacking theoretical guarantees. 
For example, surrogate models are entirely transparent but usually perform less effectively than their corresponding black-box models. This discrepancy can lead to explanations that do not accurately represent the underlying logic of the model. Instead, SHAP-based algorithms necessitate independent input features, a condition not always met in real-world applications. Other algorithms, like LIME, also have specific requirements for their correct implementation. The incorrect use of an XAI algorithm can result in misleading explanations that do not accurately address the identified legal objectives. Hence, simply employing an XAI algorithm does not guarantee adherence to the regulations discussed in this study.

Finally, our entire methodology is based on the definition of explanation from Ordinary Language Philosophy, as outlined in Section \ref{sec:methodology}. Considering alternative definitions could, therefore, introduce external threats to validity and require a different methodology.

\textbf{Intrinsic Threats.} There are possible alternative interpretations of the law's explanatory goals and high-level objectives, which our study may not fully encompass. The limited choice of case studies is another intrinsic issue, as it does not capture the complete range of nuances within the field, potentially affecting the generalisability of our results. Lastly, the list of \ac{XAI} algorithms considered in this study is not exhaustive. However, as discussed in Section \ref{sec:discussion}, our approach allows for the inclusion of new XAI algorithms and legal explanatory goals, which helps to mitigate this concern.

\section{Conclusion} \label{sec:conclusion}

This paper analysed many \ac{XAI} methods and their compliance with key EU regulations for smart biomedical devices. Significant contributions include a novel methodology for combining legal analysis, technical assessment, and a detailed categorisation of XAI methods to analyse their legal alignment. This constitutes a practical framework for selecting suitable XAI methods that help meet the explainability requirements of the \ac{GDPR}, \ac{AIA}, and \ac{MDR}.
The findings highlight the importance of XAI in meeting such demands for legal explainability. 
%
Future research should extend the case studies to various bioelectronic and biomedical devices, analysing stakeholders' perceptions of the explanations generated by XAI. 



\section*{Acknowledgements}
F. Sovrano acknowledges the support of the Swiss National Science Foundation for the SNF Project 200021\_197227. M. Lognoul acknowledges the support of the European Union H2020 research and innovation program, Grant Agreement No. 958339 (DENiM).

\bibliography{references}

\end{document}